\pdfoutput=1
\documentclass[11pt]{article}

\usepackage[final]{acl}

\usepackage{times}
\usepackage{latexsym}

\usepackage[T1]{fontenc}
\usepackage[utf8]{inputenc}

\usepackage{microtype}

\usepackage{inconsolata}

\usepackage{graphicx}

\usepackage{booktabs}
\usepackage{amsmath}
\usepackage{cleveref}
\usepackage{multirow}

\usepackage[]{todonotes}

\usepackage{xurl}

\title{Dialogue Ontology Relation Extraction via Constrained Chain-of-Thought Decoding}

\author{Renato Vukovic,
  David Arps,
  Carel van Niekerk,
  Benjamin Matthias Ruppik,\\
  {\bf Hsien-Chin Lin,}
  {\bf Michael Heck,}
  {\bf Milica Gašić} \\
  Heinrich Heine University Düsseldorf \\
  \texttt{\{renato.vukovic, david.arps, niekerk, ruppik, linh, heckmi, gasic\}@hhu.de}}

\begin{document}
\maketitle

\begin{abstract}
State-of-the-art task-oriented dialogue systems typically rely on task-specific ontologies for fulfilling user queries.
The majority of task-oriented dialogue data, such as customer service recordings, comes without ontology and annotation.
Such ontologies are normally built manually, limiting the application of specialised systems. 
Dialogue ontology construction is an approach for automating that process and typically consists of two steps: term extraction and relation extraction. 
In this work, we focus on relation extraction in a transfer learning set-up.
To improve the generalisation, we propose an extension to the decoding mechanism of large language models. 
We adapt Chain-of-Thought (CoT) decoding, recently developed for reasoning problems, to generative relation extraction.
Here, we generate multiple branches in the decoding space and select the relations based on a confidence threshold.
By constraining the decoding to ontology terms and relations, we aim to decrease the risk of hallucination.
We conduct extensive experimentation on two widely used datasets and find improvements in performance on target ontology for source fine-tuned and one-shot prompted large language models.\footnote{\scriptsize The code is available under \url{https://gitlab.cs.uni-duesseldorf.de/general/dsml/dialogue-ontology-relation-extraction-via-constrained-chain-of-thought-decoding}}

\end{abstract}

\section{Introduction}

\begin{figure*}[t]
    \centering
    \includegraphics[width=0.95\linewidth]{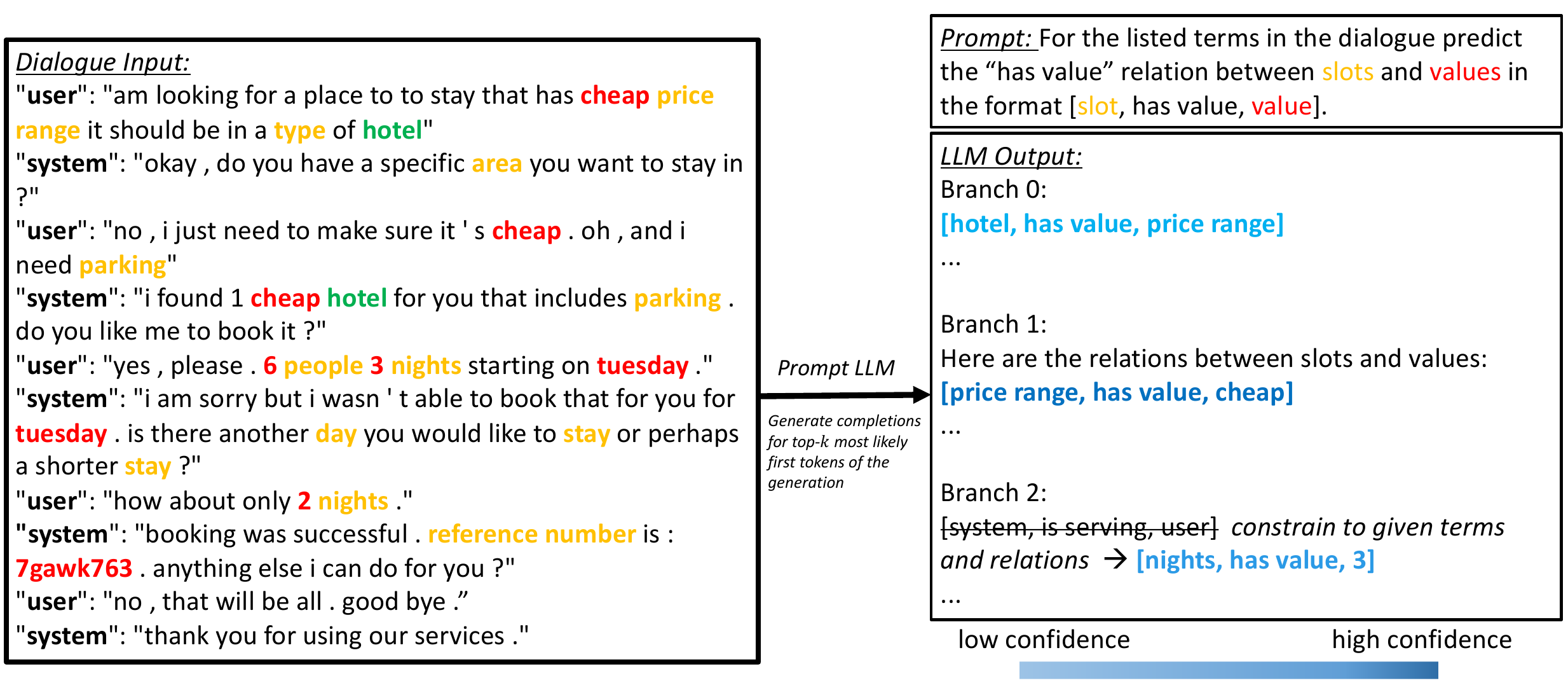}
    \caption{
        Example of constrained CoT-decoding for dialogue ontology extraction for a dialogue from MultiWOZ 2.1~\cite{eric-etal-2020-multiwoz}. 
        Domains are highlighted in green, slots in yellow and values in red. 
        Branch 0 predicts an incorrect relation (\textit{hotel} misclassified as slot) with lower confidence. 
        Branch 1 has the highest confidence in the relation prediction, which is why it is selected as the final response. 
        Also, it contains a form of reasoning that stresses the type of terms that are part of the relations to be predicted, i.e., slots and values. 
        Branch 2 visualises constrained decoding, where the prediction of terms and relations is not possible if they are not present in the input.
        \label{fig:main}
    }
\end{figure*}

State-of-the-art task-oriented dialogue (TOD) systems still rely on a fixed ontology to model their scope \citep{nguyen-etal-2023-slot,hudecek-dusek-2023-large,geishauser2024openhorizon}. 
A \emph{TOD ontology} comprises three levels of hierarchy: domains, slots and values.
\emph{Domains} are general topics of interest, \emph{slots} are types of information about entities in a domain, and \emph{values} are concrete instantiations of slots. 
Ontology thus forms a hierarchy: it is a directed graph where slots belong to domains and values in turn belong to slots. 
Note that slots can be shared across domains, and so can values.
An ontology is typically a prerequisite for generating API calls that access the underlying databases for entity retrieval. 
Further, the ontology defines the dialogue state, which is tracked by the system to determine the next actions given the evolving discourse.

The dependency on an ontology poses a significant challenge in transferring existing TOD systems to new domains and use cases. 
Although ontology-agnostic approaches do exist, their transfer capabilities are limited and their performance remains sub-par on novel data~\cite{heck-etal-2022-robust}.

Large quantities of domain-specific TOD data, e.g. customer service recordings, are frequently available, but tend to come without annotation, rendering direct use for system development difficult~\cite{BRUSCO2023101462}.
Manual labelling is error-prone, does not scale well and quickly becomes prohibitively expensive~\cite{eric-etal-2020-multiwoz,van-niekerk-etal-2020-knowing,rosenbaum-etal-2022-linguist,gung-etal-2023-intent,vanniekerk2024camell}.
Despite topical or domain mismatch, existing annotated datasets may provide information about TOD that can be leveraged to harness new data. 
For this reason, we are interested in utilising existing labelled TOD datasets to automatically generate a full ontology for new, yet-unlabelled, data.

Automatic dialogue \emph{ontology construction} typically consists of two steps, dialogue term extraction~\citep{vukovic-etal-2022-dialogue-local,ruppik-etal-2024-local} and hierarchy establishment.
Although hierarchy establishment is often done via clustering~\cite{hudecek-etal-2021-discovering,yu-etal-2022-unsupervised} we approach it via \emph{relation extraction} (RE), which is more similar to common information extraction pipelines~\cite{genest2022promptore,xu2023large}. 
We call this task \emph{dialogue ontology relation extraction} (DORE). 
A hierarchy is established by inferring in which level extracted terms lie, and by connecting terms across levels.

Although large language models (LLMs) have demonstrated considerable task transfer abilities~\cite{brown2020gpt3,ouyang2022instructgpt}, they still lack behind specialised systems in TOD modelling when appropriate training data is available~\cite{heck-etal-2023-chatgpt-local,hudecek-dusek-2023-large,feng-etal-2023-chatter}.

In this work, we assume that some labelled out-of-domain source dialogue data is available to facilitate transfer learning. 
We examine two strategies of providing source data to an instruction-tuned LLM; 1) as one-shot examples in the prompt, and 2) as data for an additional round of supervised fine-tuning. 
We establish a challenging transfer setup by conducting experiments on two well-established medium to large scale multi-domain task-oriented dialogue benchmark datasets: MultiWOZ 2.1~\cite{budzianowski-etal-2018-multiwoz,eric-etal-2020-multiwoz} and the Schema-Guided Dialogue \cite[SGD;][]{rastogi2020towards} dataset.
Since our focus is solely on DORE, we assume that the results of the first step of ontology construction, namely term extraction, are provided.

We propose to improve the decoding mechanism of an LLM in order to better leverage task-specific knowledge.
Concretely, we constrain the generation to terms and relation types given in the model input to force the model to consider terms from the target data and output the desired format.
We further adapt \emph{chain-of-thought (CoT) decoding}~\cite{wang2024chain}, which was recently proposed for logical reasoning, for DORE.
Traditionally, CoT methods prompt or train the model to generate reasoning paths before giving the final answer \cite{wei2022chain,kim-etal-2023-cot}.
CoT-decoding on the other hand exploits the observation that the presence of CoT-paths is correlated with higher confidence in the predicted answer in logical reasoning.
We extend CoT-decoding to DORE by selecting the final answer based on the confidence of predicted relations in multiple generated model answer branches.
Our final proposal, \emph{constrained CoT-decoding for dialogue ontology extraction}, is the combination of our CoT-decoding approach to RE with constrained decoding, see Fig.~\ref{fig:main}.
Empirically, this new decoding mechanism significantly outperforms both source one-shot and source fine-tuned baselines on the target data.
Our contributions are as follows:
\begin{itemize}
    \item We propose to induce an ontological hierarchy by accumulating ontology relation predictions from the dialogues in a TOD dataset.
    \item To the best of our knowledge, we are the first to apply \emph{CoT-decoding} to dialogue ontology relation extraction.
    \item We develop an extension, called \emph{constrained CoT-decoding}, for multi-relation extraction from task-oriented dialogues.
    \item Constrained CoT-decoding significantly improves the quality of relation predictions on the target dataset for both source one-shot and source fine-tuned baselines.
\end{itemize}

% % % % % % % % % % % % % % % %
\section{Related Work\label{sec:related-work}}

% % % % % %
\paragraph{Dialogue Ontology Construction}
We divide dialogue ontology construction into term extraction and relation extraction. 
\citet{vukovic-etal-2022-dialogue-local} improve out-of-domain generalisation of a dialogue term extraction model by making use of topological properties of the language model embedding space.
\citet{nguyen-etal-2023-slot} improve phrasal segmentation of ontology terms via language model probing and contrastive learning.
Since we evaluate the hierarchy on a global level based on relations, our approach is not directly comparable to clustering-based approaches such as \citet{hudecek-etal-2021-discovering,yu-etal-2022-unsupervised}.
In contrast to these methods, we view hierarchy establishment in isolation as a relation extraction task.

\citet{yu-etal-2020-dialogue} present DialogRE, a popular dataset for RE in short chit-chat dialogues. 
Closest to our approach, 
\citet{albalak-etal-2022-rex} jointly optimise RE and explanation generation to improve performance with a model-agnostic framework.
\citet{xu-chen-2023-zero} propose a zero-shot approach for extracting trigger words for dialogue relation extraction on DialogRE. 
However, these works focus on chit-chat dialogues, which do not include ontology relations.

% % % %
\paragraph{Relation Extraction with LLMs}

LLMs show promising transfer capabilities out of the box~\cite{laskar-etal-2023-systematic}. 
Direct application to our task however is not promising, as it has been shown that aligned LLMs such as ChatGPT~\cite{chatgpt_blog_openai_2023} do not perform well on extracting multiple relations at once~\cite{lilong2024autore}. 
This shortcoming has been linked to the influence of pre-training data distribution on downstream task performance~\citep{McCoy2023EmbersOA}. 
RE data in particular amounts to a mere 0.5\% of instruction-tuning datasets, and is hardly utilised for model selection~\citep{wang-etal-2022-super,zhang-etal-2023-aligning}. 

Traditionally, RE is performed in a pairwise manner~\cite{zhang-etal-2023-aligning}, resulting in quadratic complexity given the number of terms. 
This becomes intractable for generative LLMs when querying the LLM separately for each pair of terms. 
Alternatively, one may extract all relations present in a given input with a single LLM query, as is common in multi-relation extraction tasks such as document-level RE. For example, \citet{lilong2024autore} extract relations by either predicting relations directly, or first predicting possible head entities in a document. 
\citet{zhang-etal-2023-aligning} align LLMs for zero-shot RE by transforming RE into a question answering (QA) task, which is more frequent in the instruction-tuning data. 

\begin{table*}[t]
    \footnotesize
    \centering
    \begin{tabular}{ccc}
        \toprule
        \textbf{Relation} & \textbf{Verbaliser} & \textbf{Example}  
        \\ 
        \midrule
        Domain-Slot & [Domain, has slot, Slot] & [hotel, has slot, price range] \\
        Slot-Value & [Slot, has value, Value] & [price range, has value, cheap] \\
        Value-Domain & [Value, has domain, Domain] & [cheap, has domain, hotel] \\
        Equivalence & [Term1, refers to same concept as, Term2] & [cheap, refers to same concept as, low budget] \\
        \bottomrule
    \end{tabular}
    \caption{
        \label{tab:relation_definitions}
        Hierarchical dialogue ontology relation task definition with examples.
    }
\end{table*}

\paragraph{Constrained Decoding} 
Constrained decoding limits the tokens that can be generated. 
It is typically applied to LLMs to improve downstream task performance, reduce hallucination and ensure certain output formats.
\citet{bogoychev-chen-2023-terminology} constrain decoding for translation to ensure that certain terminology is used.
\citet{Roy2024FLAP} use constrained decoding with a lookahead heuristic to speed up adaptation of LLMs to plan generation according to a given API in TOD.
We want to force the model to use its inherent task knowledge while transferring abilities to new data.

\paragraph{Chain-of-Thought Reasoning}
LLM performance on complex reasoning tasks improves when the model generates a chain of thought (CoT). \citet{wei2022chain} include examples of multi-step reasoning in the prompt, and \citet{kojima2022large} prompt the model in a zero-shot fashion to ``think step by step''. 
Reasoning capabilities can be further enhanced via specific training on CoT-data~\cite{chung2024scaling}, or via teaching the model to reason~\cite{zelikman2022star}.
In contrast to this, we focus on eliciting model-inherent reasoning capabilities, without the need for specific prompts or training. As described in Sec.~\ref{sec:cot-decoding}, we leverage the fact that a top-$k$ decoding beam usually contains a CoT \cite{wang2024chain}.

% % % % % % % % % % % % % % % % % % % % % % % % % % % % % % % % % % % % % % % % % % % % % % % %
\section{Constrained Chain-of-Thought Decoding for Ontology Relation Extraction}

% % % % % % % % % % % % % % % %
\subsection{Problem Definition}

Dialogue ontology relation extraction (DORE) aims at extracting all relations between different terms in a TOD dataset.
As seen in \Cref{fig:main}, for each dialogue paired with a list of ontology terms, the output is a set of relations similar to document-level relation extraction \citep{tan-etal-2022-revisiting}. 
However, we consider the joint relation prediction set accumulated from all dialogue-level predictions, rather than the dialogue-level performance. 
In the DORE task, the model receives as input a task-oriented dialogue $D$ annotated with a list of ontology terms $T$ present in this dialogue.
The output are valid ontology relations $R_{D,T}$ between the terms, which includes predicting whether a term is a \emph{domain, slot}, or \emph{value}.
A relation is denoted by a relational triplet with a head term, the relation and a tail term.
Finally, the predicted relations for each dialogue are unified to form the final ontology relation set.

We consider 4 types of relation between ontology terms: \emph{domain-slot}, \emph{slot-value}, \emph{value-domain} and \emph{equivalent term} relations (see \Cref{tab:relation_definitions} for examples).
Here, all relations except the equivalence relation are directed relations with a head and a tail term.
\emph{Domains} are general topics, such as \textit{hotel} or \textit{restaurant}, \emph{slots} are types of information for entities in a \emph{domain}, such as \textit{price range} or \textit{area} and \emph{values} are concrete instantiations of slots, such as ``cheap'' or ``west''.
The equivalence relation connects terms from the same hierarchy level that point to the same ontological concept, e.g. “expensive” and “high-end” both represent a high price.
In the prompt and labels, we denote the relation types through different verbalisers, shown in \Cref{tab:relation_definitions}. 
Verbalisers are descriptions of task-specific labels in natural language. They align the task closer with the pre-training distribution of the LLM~\cite{schick-schutze-2021-exploiting,mosbach-etal-2023-shot}.

% \paragraph{General Definitions for Dataset Transfer}
Our hypothesis is that the general definitions of the ontology hierarchy relations enable seamless transfer to new data in order to construct a similarly structured ontology on the new data. 
% This is also possible if the target data has significantly more \emph{domains} than the source data.
Based on these relations, we focus on transferring the structural information about ontologies from a source dataset to a target dataset.
Here, we consider a one-shot and a fine-tuning approach.

% % % % % % % % % % % % % % % % % % % % %
\subsection{Chain-of-Thought Decoding\label{sec:cot-decoding}}

CoT reasoning in LLMs has demonstrated improved performance in various complex reasoning tasks (Sec.~\ref{sec:related-work}). 
The results of \citet{wang2024chain} show that LLMs inherently possess reasoning capabilities, which can be elicited without explicit prompting through \emph{Chain-of-Thought decoding}.
Concretely, they experiment on pre-trained and instruction-tuned versions of PaLM 2 \cite{anil2023palm} and Mistral-7B \cite{jiang2023mistral}. 
They observe that although the greedily decoded response might not always exhibit reasoning, one of the top-$k$ beams usually contains a CoT.
This CoT not only shows higher confidence in the answer, but also exhibits greater accuracy.
They propose to consider the top-$k$ probability tokens at the start of the predicted response.
From there, $k$ completions, called \emph{branches}, are generated, resulting in $k$-times computational complexity during inference.
The final response is chosen based on the confidence of the tokens that belong to the \emph{answer} in each branch, i.e., the average confidence of the \emph{answer tokens}.
In logical reasoning, there is only one answer in each branch, which is a number.
In that case, they identify the answer by prompting the model with “So the answer is:” at the end and match the following number to one in the preceding response.
In our case, there are multiple answers per branch, which we identify based on the fact that relations are supposed to be predicted between brackets.

% % % % % % % % % % % % % % % % % %
\paragraph{CoT-Decoding for DORE}

In this paper, we extend CoT decoding to handle the multi-answer scenario in the DORE task.
We compute the confidence of answer tokens by utilising their structure, which, in our case, involves predicting relational triplets in the format $[\textit{headterm}, \textit{relation}, \textit{tailterm}]$ and the notion of disparity. 
The disparity of a probability distribution is the difference between the probability of the most likely outcome and the next most likely outcome.
The confidence for each answer token for a given branch is measured by the average disparity of its tokens. 
Formally this is given by
\begin{equation}
        \Delta_{i, \text{a}} = \frac{1}{n} \sum_{x_t \in \text{a}} p(x_t^{\text{top}} \mid x_{<t}) -  p(x_t^{\text{next}} \mid x_{<t}),
\end{equation}
where $a$ is an answer (in our case the triplet), $i$ is a branch, $x_t$ are the answer tokens belonging to the answer in branch $i$,
$x_t^{\text{top}}$ is the most likely token on position $t$ and $x_t^{\text{next}}$ the next most likely token on position $t$.
$x_{<t}$ are the tokens in branch $i$ on positions preceding $t$, i.e. the context so far.

In DORE, answer tokens are those that form terms and relations in the predicted relational triplets, which means there are three disparities per relation.
This approach relies on detecting answer tokens in a generated response for confidence estimation, and we leave an extension to arbitrary answer structures to future work.
The resulting triplet disparities are denoted as $\Delta_a = [\Delta_h, \Delta_r, \Delta_t]$.
We explored mean, median, maximum, and minimum as aggregation strategies for relational triplet mentions, finding that all of them lead to similar results.
For simplicity, we choose the mean to aggregate the disparity for a relational triplet in branch $i$, i.e. $\Delta_{i,a} = \frac{1}{3} (\Delta_{h,i} + \Delta_{r,i} + \Delta_{t,i})$.

We select the branch with the highest average disparity over the relations predicted in each branch to get the final set of relation predictions for a dialogue.
The average disparity for branch $i$ is given by 
\begin{equation}
    \overline{\Delta}_{i} = \frac{1}{n_{a,i}} \sum_{a \in R_{i}} \Delta_{i,a}
\end{equation}
where $a$ is a relational triplet, $R_{i}$ is the set of relations and $n_{a,i}$ is the number of relations in branch $i$.
The final set of predicted relations is then given by
\begin{equation}
    R_{\overline{\Delta}_{\max}} 
    = 
    \{
        R_{i} 
        \mid 
        i = \text{argmax} \{
            \overline{\Delta}_{0}, \ldots, \overline{\Delta}_{k} 
        \}
    \}
\end{equation}

We also experiment with a confidence threshold based approach for relation selection.
Here, the average disparity of a relation is computed across occurrences in different branches:
\begin{equation}
    \widetilde{\Delta}_{a} 
    = 
    \frac{1}{n_a} \sum_{i \in \{1,...,k\}} \Delta_{i,a},
\end{equation}
where $\Delta_{i,a}$ is the disparity of the answer $a$ in the $i$-th branch and $n_a$ is the number of occurrences of $a$ across the different branches.
The final set of predicted relations $R_{\Delta > \Delta_{\text{threshold}}}$ is then
\begin{equation}
    R_{\Delta > \Delta_{\text{threshold}}} 
    = 
    \{a \mid \widetilde{\Delta}_{a} > \Delta_{\text{threshold}} \}
\end{equation}

\subsection{Constrained Decoding}
We constrain the generation of the relation terms and relation types if the beginning of a relational triplet is predicted to ensure the structure and mitigate term and relation hallucination (see \Cref{fig:main}).
This means for a relational triplet, $[h, r, t]$, we ensure that $h,t \in T$ and $r \in R$, where $T$ is the set of terms for the current dialogue and $R$ is the set of relation types given in the prompt.
Note that we only constrain the generation when an opening bracket is predicted by the model, and resume to non-constrained generation after the generated relational triplet.

% % % % % % % % % % % % % % % %
\section{Experiments}

% % % % % % % % % % % % % % % %
\subsection{Experimental Setup}\label{ss:setup}

We utilise the open-source Gemma 2B~\citep{team2024gemma} instruction-tuned model with context size of 4096  for all experiments.
In CoT-decoding we set $k=5$.
For a more thorough analysis of the impact of $k$ in CoT decoding, resort to \citet{wang2024chain}.
We always branch at the first token; branching at later tokens did not show improvements.
For all CoT-decoding experiments, we select the relations from the branch with the highest disparity, as the threshold based method works worse and also adds a new hyperparameter.
In the one-shot prompts, we use a combination of an instruction with simple natural language with a preceding example~\cite{brown2020gpt3,sahoo2024systematic}.
For fine-tuning, we remove the example from the prompt.

% % % % % % % % % % %
\paragraph{Datasets}

For the source dataset, we employ the MultiWOZ 2.1 dataset \citep{eric-etal-2020-multiwoz}.
It has 7 domains and over 10,000 dialogues.
We use the training set for training and select from it one random dialogue with relation annotation as one-shot exemplar.
The target dataset is the schema-guided dialogue (SGD) dataset \citep{rastogi2020towards}.
It comprises more than 20,000 dialogues and 20 domains.
We use the SGD test split for evaluation in the main results, which contains 4,201 dialogues and 18 domains.
%We use the first 1,000 dialogues for easier reproducibility.
%These contain 15 of 18 domains from the whole test set in a similar distribution to the whole test set. Consequently, our results can be considered representative.% for the whole test set, with only small differences in domain distribution.
%This $\frac{1}{4}$ of the dialogues of the total SGD test set 
%Our test set portion contains 35\% of ontology relations from the full test set.
%We also checked the performance difference between whole test set and the considered subset for one MultiWOZ fine-tuned seed, which was less than 0.01 in F1 score.
%In the subset, there are 167 domain-slot relations, 3,249 slot-value relations, 3,274 value-domain relations and 186 equivalence relations.
In the test set, there are 134 domain-slot relations, 6,162 slot-value relations, 8,233 value-domain relations and 330 equivalence relations.
It is worth noting the SGD test set contains dialogues from different domains than the SGD training set, as well as a significant amount of unseen ontology relations.
We use ConvLab-3~\cite{zhu-etal-2023-convlab-local} for loading all the datasets.
%info for the whole test set
%Here, there are 8,233 value-domain relations, 6,162 slot-value relations, 134 domain-slot relations and 330 equivalence relations.
%

\paragraph{Training}

\begin{table*}[t]
    \centering
    \resizebox{0.78\linewidth}{!}{%
    \begin{tabular}{l|ccc}
        \toprule
        \textbf{Approach} & \textbf{F1-Score} & \textbf{Precision} & \textbf{Recall} \\
        \midrule
        \multicolumn{4}{c}{
            \textit{One-shot example from MultiWOZ}
        } \\
        % \midrule
        % {One-shot example from MWOZ} & 0.021 & 0.051 & 0.013 \\ 
        \midrule
        {\textit{Baseline}: Separate relation prediction} & 7.4 & 8.8 & 6.4 \\ 
        {\hspace{0.5cm}+ constrained decoding} & \textbf{8.5*} & 5.7  & \textbf{17.3*}$\mathbf{\ddagger}$ \\ 
        {\hspace{0.5cm}+ CoT decoding} & \textbf{9.2*} & 8.8 & \textbf{9.6*} \\ 
        {\hspace{0.5cm}+ constrained CoT decoding} & \textbf{9.2*} & 6.4 & \textbf{15.9*} \\ 
        \midrule
        \multicolumn{4}{c}{
            \textit{Fine-tuning on MultiWOZ}
        } \\
        \midrule
        {\textit{Baseline}: Fine-tuning on MultiWOZ} & 10.9 & 6.8 & 28.8 \\ 
        {\hspace{0.5cm}+ constrained decoding} & \textbf{12.0}$\mathbf{\dagger}$ & 7.4 & 32.3$\mathbf{\ddagger}$ \\ 
        {\hspace{0.5cm}+ CoT decoding} & 10.6 & 7.6 & 17.6 \\ 
        {\hspace{0.5cm}+ constrained CoT decoding} & \textbf{13.7}$\mathbf{\dagger}\mathbf{\ddagger}$ & \textbf{9.8}$\mathbf{\dagger}$ & 23.0 $\mathbf{\ddagger}$ \\ 
        \midrule
        \multicolumn{4}{c}{
            \textit{Upper Bounds using SGD Data}
        } \\
        \midrule
        % {One-shot example from SGD} & 0.016 & 0.067 & 0.009 \\ 
        {One-shot example from SGD + separate relation prediction} & 12.9 & 10.7 & 16.4 \\ 
        %{+ constrained Decoding} & 0.133 & 0.114 & 0.161  \\ 
        %{+ CoT Decoding} & 0.117 & 0.095 & 0.149 \\ 
        %{+ constrained CoT Decoding} & 0.112 & 0.089 & 0.151 \\ 
        {Fine-tuning on SGD} & 37.3 & 27.9 & 57.2 \\ 
        %{+ constrained Decoding} & 36.7 & 25.8 & 63.8 \\ 
        %{+ CoT Decoding} & 1.7 & 19.5 & 0.9 \\ 
        %{\hspace{0.5cm} \textcolor{gray}{+ constrained CoT Decoding}} & \textcolor{gray}{14.6} & \textcolor{gray}{14.8} & \textcolor{gray}{14.4} \\ 
        \bottomrule
    \end{tabular}%
    }
    \caption{
        \label{tab:sgd_test_gemma_results}
        Ontology Relation Prediction Results on the SGD test set. 
        Results that are statistically significant over the baseline are highlighted in \textbf{bold}. 
        Additionally, significant results based on dialogue-level evaluation for one-shot prompts are marked with \textbf{*}. 
        Significant results for fine-tuned models, evaluated globally based on five random seeds, are marked with $\mathbf{\dagger}$. 
        Significant improvement over the one-shot model from the SGD upper bound on dialogue-level is marked with $\mathbf{\ddagger}$. 
        All significance tests are performed at a $5\%$ level of significance.
    }
\end{table*}

For both fine-tuning and one-shot prompting, we utilise the original Gemma prompt template~\cite{team2024gemma}.
For training, we utilise Low-rank adaptation~\cite[LoRA,][]{hu2022lora} with the default parameters in the peft library~\cite{peft}.
We train the model on a single NVIDIA RTX8000 GPU and do inference with all models on one NVIDIA RTX6000 GPU. 

We only consider a one-shot approach due to context size constraints, as the relational triplets in the exemplars contain brackets.
Brackets are considered individual tokens, increasing the number of tokens significantly. 
Because of this a maximum of three exemplars fits in the context size, which do not improve performance however, while increasing computational complexity. 
In the one-shot approach, we predict each relation type separately, since we found that the LLM struggles with jointly predicting all relation types. 
We also experimented with a zero-shot approach that performs significantly worse than one-shot.

We fine-tune the LLM via pattern-based fine-tuning~\cite{schick-schutze-2021-exploiting,ma-etal-2023-prompt-local} with a prompt for all relation types on the MultiWOZ training split.
We consider two upper bounds: an LLM trained on the SGD training split and a model utilising a one-shot exemplar from SGD.

% % % % % % % % % % % % %
\subsection{Evaluation}

In evaluation, we only consider relations within dialogues in the ground truth, i.e., both terms of a relation occur in the same dialogue.
Relations from equivalent terms to other terms have to be found at least once.
If $[\textit{term}_1, \textit{refers to same concept as}, \textit{term}_2] \in R_{\text{groundtruth}}$, then $[\textit{term}_1, r, t] = [\textit{term}_2, r, t]$, where $R_{\text{groundtruth}}$ is the set of ground truth relations, $r \neq \text{`refers to same concept as'}$ is another relation type and $t \in T$ is a third related term.
E.g., the relations [price range, has value, high-end] and [price range, has value, expensive] are equivalent, since [expensive, refers to the same concept as, high-end].
Thus, the prediction of the former relation counts as a prediction for the latter and vice versa.

To compute the global micro F1 score, we compare the accumulated set of relations predicted from all the dialogues with the ground truth ontology relations.
Note that we only consider exactly matching terms in relations to be correct.

For significance tests on the one-shot prompted models, we employ a pairwise $t$-test on dialogue level.
For fine-tuned models, we use 5 random seeds for training and an independent $t$-test.

% % % % % % % % % % %
\subsection{Results}

\Cref{tab:sgd_test_gemma_results} shows the full results on the target test set, see \Cref{sec:appendix:relation_type_results} for results for each relation type.

% % % % % %
\paragraph{Source One-Shot Approach}

We found that when predicting all relations at once in a one-shot fashion the model is completely unable to fulfil the task, so we resort to predicting one relation at a time.
The one-shot approach is mainly improved through constrained decoding, although the combination with CoT-decoding is also significantly better than the baseline.
Note that the source one-shot model is able to get closer to the performance of a model with a one-shot example from the target data with constrained CoT-decoding. 
% % % % % %
\paragraph{Source Fine-tuning Approach}
\label{paragraph:source-fine}
For the source fine-tuned model, \emph{constrained CoT-decoding} leads to significant improvements over the baseline. 
Furthermore, it significantly outperforms a model using a one-shot exemplar from the target data on all metrics.
Constraining CoT-decoding helps performance, since the constraints mitigate overconfidence on the source data after fine-tuning.

Interestingly, although the target fine-tuned model is the best model, it is not able to find all relations on the test set. 
As mentioned in Section~\ref{ss:setup}, the SGD test set contains domains different to the SGD training set, which makes this task particularly difficult. 
In contrast to the excellent performance of LLMs on a variety of tasks, there is a lot of room for improvement on this task.

% \begin{table}[t]
%     \centering
%     \resizebox{0.8\columnwidth}{!}{%
%     \begin{tabular}{c|c|c|c}
%         Aggregation & F1 Score & Precision & Recall \\ \midrule
%         \multicolumn{4}{c}{MWOZ One-shot constrained CoT decoding} \\ \midrule
%         Mean & 11.2 & 8.3 & 17.4 \\
%         Max & 11.3 & 8.3 & 17.7 \\
%         Min & 11.3 & 8.4 & 17.6 \\
%         Median & 11.9 & 8.8 & 18.2 \\ \midrule
%         \multicolumn{4}{c}{MWOZ Fine-tuned constrained CoT decoding} \\ \midrule
%         Mean & 19.6 & 14.1 & 32.1 \\
%         Max & 19.5 & 14.1 & 31.3 \\
%         Min & 19.6 & 14.2 & 31.8 \\
%         Median & 19.6 & 14.1 & 31.9 \\
%     \end{tabular}%
%     }
%     \caption{F1 score of different aggregation strategies on the SGD test set. Use the highest disparity branch relation selection method here for visualisation purposes.}
%     \label{tab:different_aggregation_strategies}
% \end{table}

% \paragraph{Impact of Aggregation Function}

% In \Cref{tab:different_aggregation_strategies} we see that the aggregation functions of relation triplet disparity do not have a significant impact on the resulting performance of the highest disparity selection method.
% Because of that, we resort to the mean of the relation triplet disparities for aggregation.

% % % % % % % % % % % % % % % % % %
\subsection{Calibration Analysis}

\begin{figure}
    \centering
    \includegraphics[width=0.95\linewidth]{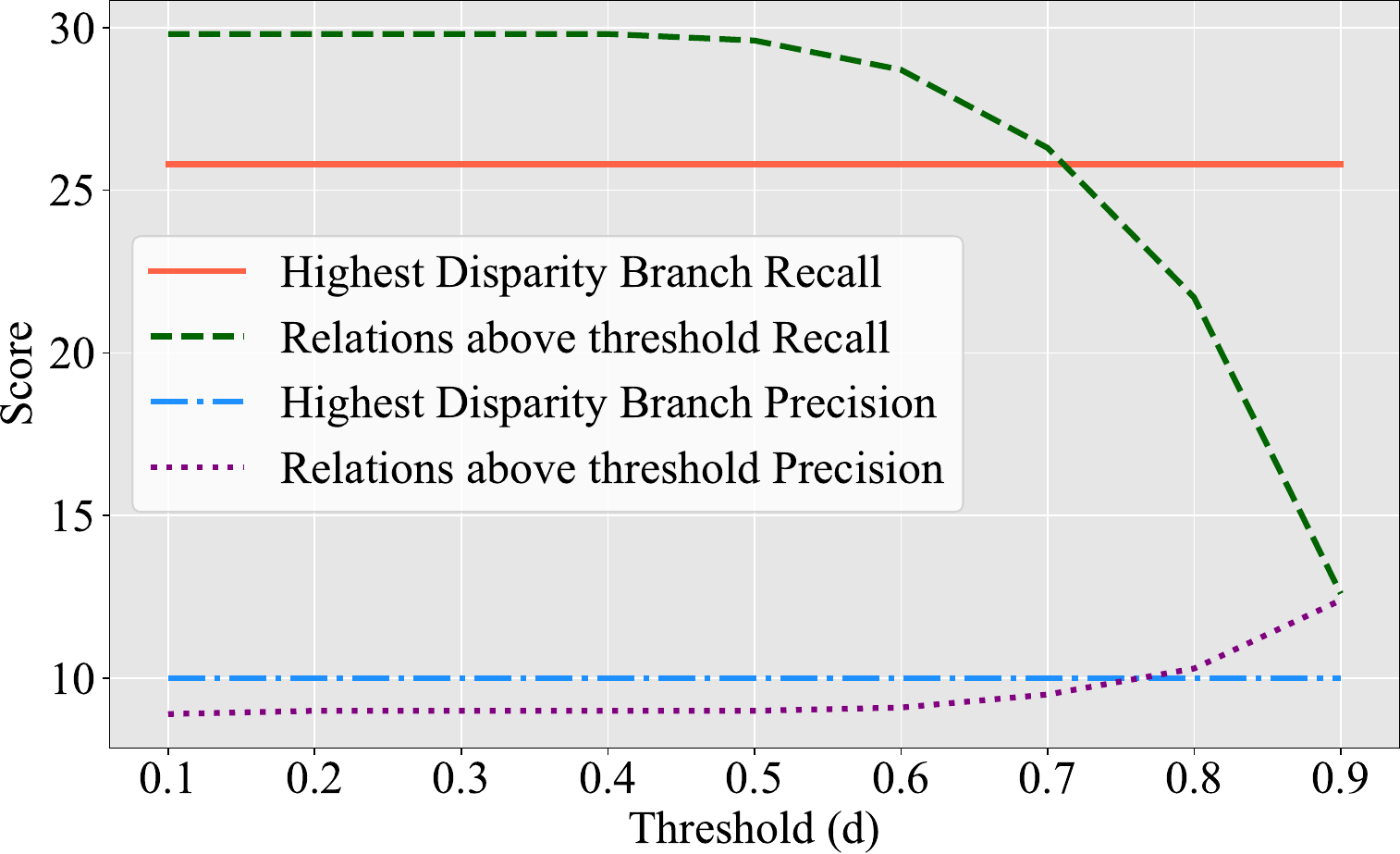}
    \caption{
        \label{fig:different_thresholds}
        Different relation confidence thresholds across branches compared to the \emph{highest disparity branch approach} for mean aggregation. 
        Displayed are recall and precision for the MWOZ fine-tuned constrained CoT decoding model.
    }
\end{figure}

In \Cref{fig:different_thresholds}, we see that an absolute confidence threshold is not as meaningful and adds the problem of choosing the correct threshold as hyperparameter.
Moreover, a high threshold leads to only a small increase in precision, while losing a significant amount of recall.
Our results are in line with recent findings about instruction-tuned LLMs~\cite{kapoor-etal-2024-calibration} being overconfident.
We find that the model's confidence on predicted relations is generally on a high level, indicating overconfidence, as the significant changes in performance happen at high confidence thresholds.
For lower thresholds, the performance remains unchanged, as most confidences are quite high and hence the set of predicted relations stays the same. Although this shows that the thresholds are less meaningful, the relative confidence of the branches is meaningful, since choosing the highest disparity branch leads to good performance.

% % % % % % % % % % % % % % % % % % % % % % % % % % % % % % % % % % %
\subsection{How useful are predictions from the additional branches?}

\begin{figure}[t]
    \centering
    \includegraphics[width=0.95\linewidth]{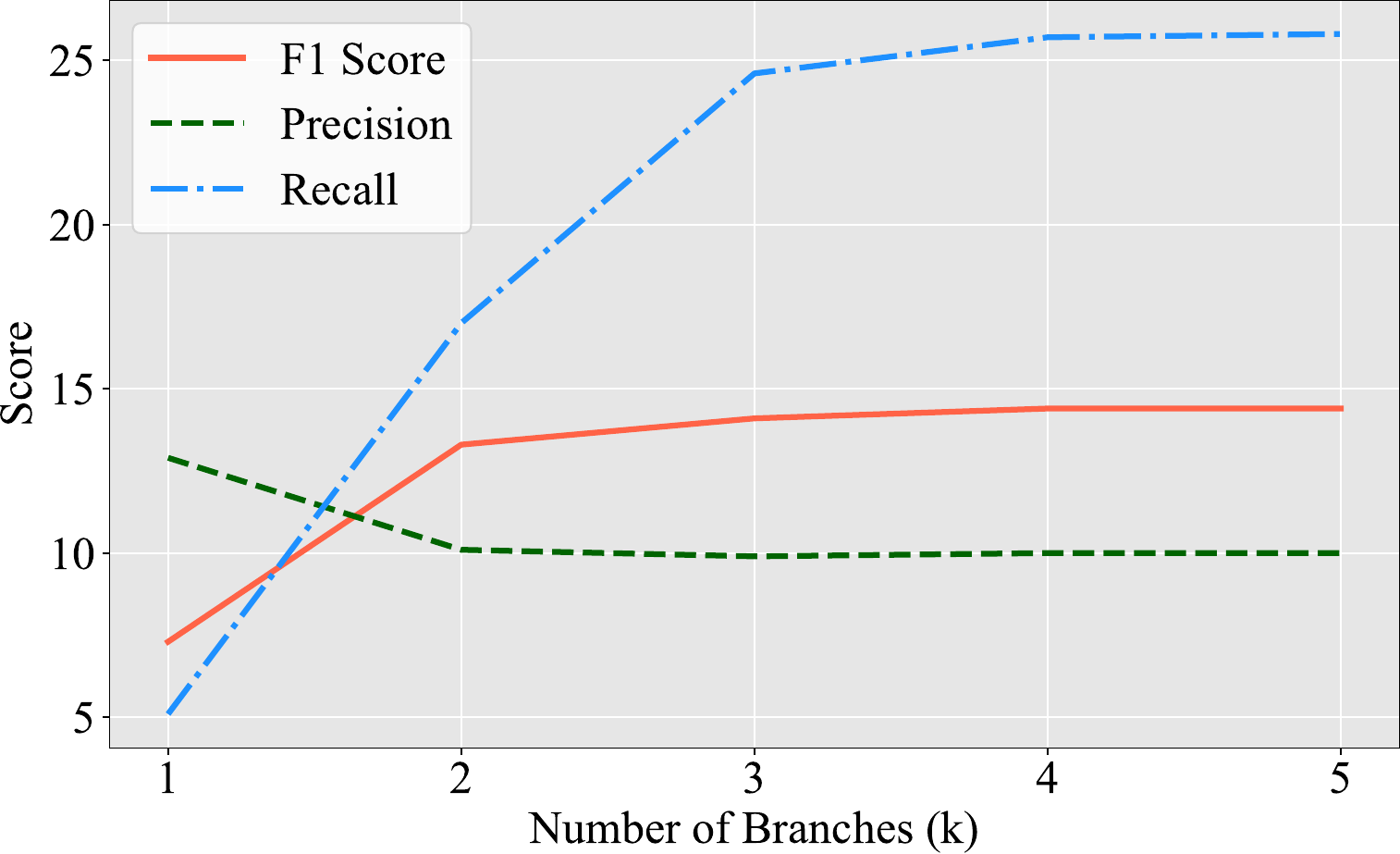}
    \caption{
        \label{fig:diff_k_constrained_cot_finetuned}
         MWOZ fine-tuned constrained CoT decoding model with different choices for the number of branches $k$ for the \emph{highest disparity branch method}. Shown are SGD test set F1 score, precision, and recall.
    }
\end{figure}

In line with the findings from \citet{wang2024chain}, we find that for the instruction-tuned Gemma model, the gain in performance can be mainly attributed to the first additional branch (see \Cref{fig:diff_k_constrained_cot_finetuned}).
While the F1 score is increased slightly up to $k=3$, the jump in recall from $k=2$ to $k=3$ is more significant.
This also shows that the branches from lower ranked first tokens lead to responses with higher total confidence across the relation predicted in the respective branch, which is why they are chosen in the highest disparity branch selection method.

% % % % % % % % % % % % % % % % % %
\subsection{Qualitative Analysis}

\begin{figure}%[ht]
    \centering 
    \includegraphics[width=0.95\linewidth]{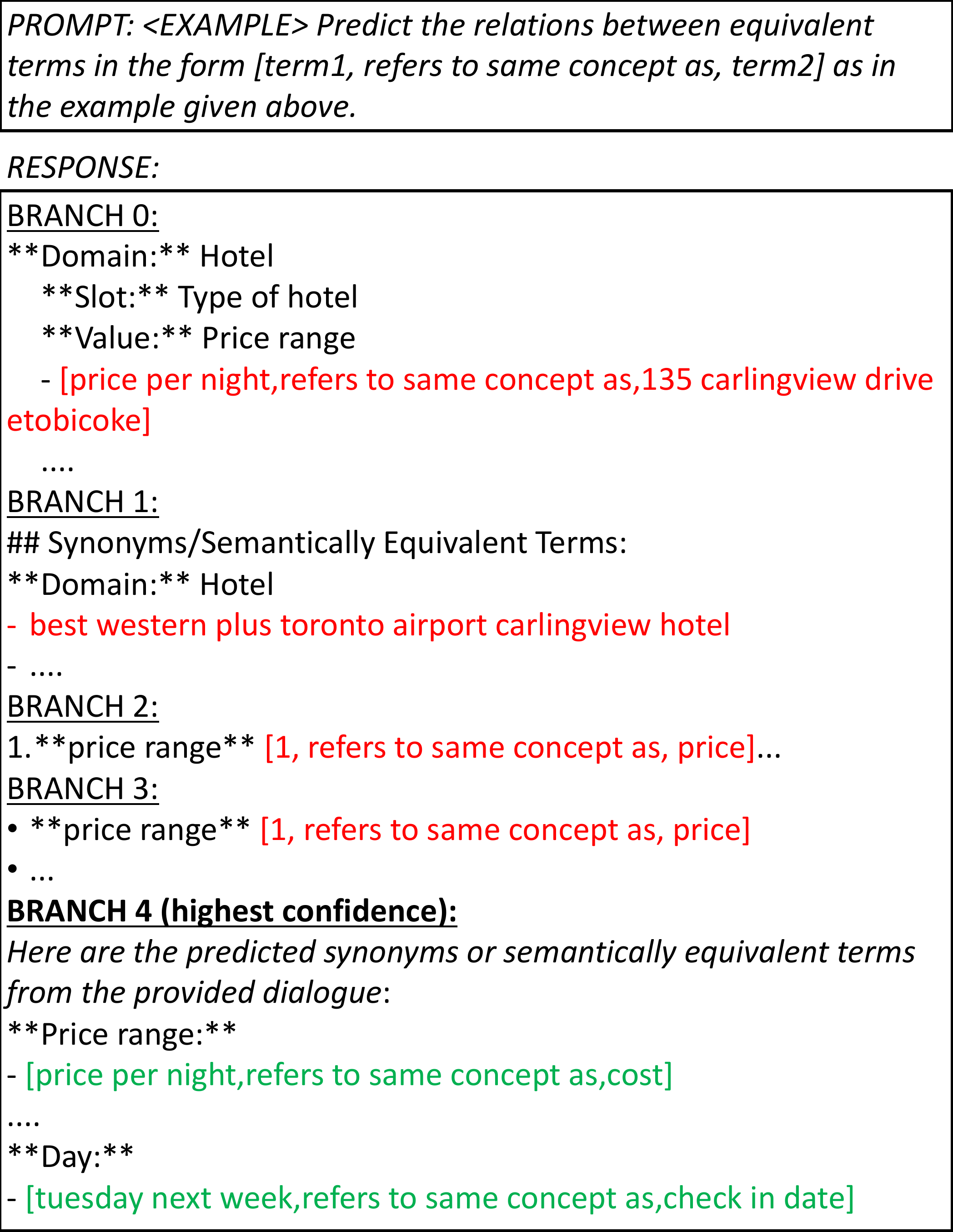}
    \caption{
        \label{fig:branch_reasoning_example}
        Example of constrained CoT decoding for one-shot equivalence relation prediction. 
        Branch 4 adds focus on the relation type. 
        It has the highest average confidence in the predicted relations and is chosen as the final response.
        Some response parts are left out for illustration purposes (“....”).
    }
\end{figure}

\citet{wang2024chain} found that LLMs struggle to generate CoTs for less frequent tasks in the pre-training data.
In our analysis, we found that higher confidence completions often follow a recap of the type of terms and relations that should be predicted.
Illustrated in \Cref{fig:branch_reasoning_example} is an example of a response to the one-shot equivalence prediction prompt with constrained CoT-decoding (see \Cref{sec:appendix:only_eq_examples} for completions of the other decoding approaches).
Here, branches 0 and 1 contain a repetition of the information given in the prompt.
In branch 1 however, the focus on the equivalence mentioned in the prompt is followed by a response that does not use the proper format for the answer to be parsed correctly.
The last branch has the highest confidence and is chosen ultimately.
Here, the focus on the task relation and the provided dialogue is part of the generated introduction to the response.

As seen in \Cref{tab:sgd_mwoz_tuned_qualitative_example}, for the fine-tuned models, there are no such reasonings observed, although the constrained CoT decoding significantly improves performance.
We hypothesise that the additional branches together with the constraints in decoding force the model to use task knowledge from fine-tuning, rather than what it has learned about the source data distribution. 
This can be observed when comparing CoT-decoding with constrained CoT-decoding, where the unconstrained version mainly generates terms it has seen on MultiWOZ, such as the ``reference number'' slot that is not present in SGD.
The constrained version on the other hand forces the model to use task knowledge instead of distributional knowledge, leading to a much better coverage of the terms mentioned in the dataset, if the correct branch is chosen based on confidence.
When observing completions to other dialogues, we found that the qualitatively best branches are not necessarily those with the highest confidence, indicating that a more sophisticated branch selection strategy might boost performance further.
We leave such an improvement to future work.
When comparing constrained decoding with vanilla greedy decoding, it becomes apparent that constraining the generation greatly improves the output structure and the utilisation of mentioned terms in the target dataset.

\begin{table*}[t]
    \footnotesize
    \centering
    %\resizebox{0.78\linewidth}{!}{%
    \begin{tabular}{p{0.13\linewidth}|p{0.8\linewidth}}
        \toprule
        Approach & Response \\
        \midrule
        Baseline & - [price per night, has domain], [best western plus toronto north york hotel \& suites, has domain], '\$ 63', 'hotel', 'has slot', 'has domain', 'hotel' ] nessunize "\$ 63" \$ 117' - [best western plus toronto north york hotel \& suites, has domain], 'has slot', 'hotel', 'has domain', 'attraction' ] - [hotel, has slot],'refers to same concept as','stay' ] - [hotel, has slot], 'has domain', 'hotel' ] ... \\
        \midrule
        Constrained Decoding & - [price per night,has value,\$ 117] - [hotel,has slot,price per night] - [best western plus toronto north york hotel, has domain, hotel] - [hotel,has slot,address] - [best western plus toronto north york hotel, has domain, restaurant]... [hotel,has slot,date] ... \\
        \midrule
        CoT-Decoding & [1, has domain, hotel] [best western plus toronto north york hotel \& suites, has domain, hotel] [1, has domain, address] [best western premier toronto airport carlingview hotel, has domain, hotel] [hotel, has slot, day] [best western plus toronto north york hotel, has domain, hotel] [hotel, has slot, name] [hotel, has slot, room] [hotel, has slot, area] [hotel, has slot, reference number] ... \\
        \midrule
        Constrained CoT-Decoding & [1, has domain, hotel] ... [hotel,has slot,price per night] [best western plus toronto north york hotel, has domain, restaurant] [best western plus toronto north york hotel, has domain, hotel] [hotel,has slot,address] [hotel,has slot,check in date] [hotel,has slot,street address] [hotel,has slot,date] [hotel,has slot,tuesday next week] [hotel,has slot,check in date] ...  \\
        
        \bottomrule
    \end{tabular}%
    %}
    \caption{
        \label{tab:sgd_mwoz_tuned_qualitative_example}
        MWOZ fine-tuned model example response excerpts for the different decodings on SDG test dialogue 100. 
        For CoT-decoding methods, only the chosen branch is displayed.
    }
\end{table*}

% % % % % % % % % % % % % % % % % %
\section{Discussion}

Although the performance of the fine-tuned model is improved by constrained CoT-decoding, it is not clear where the improvement comes from based on qualitative analysis alone, as this model generates no reasoning paths.
The workings and explainability of constrained CoT-decoding for fine-tuned models have not been investigated yet, but are relevant future research directions.

Our results imply that for tasks that are not frequently used in instruction-tuning data, it can be useful to utilise existing TOD data for training an LLM rather than annotating a few examples on the target data for the DORE task (see \Cref{paragraph:source-fine}).
Due to the length of examples in the DORE task, the amount of annotated examples that can fit in the prompt is highly limited, aggravating the applicability of few-shot approaches.

The results strengthen the finding that annotated data should be used if available~\cite{zhu-etal-2023-weaker,feng-etal-2024-infusing}.
Constrained CoT-decoding only improves performance on the target data, where task knowledge is more useful than distributional knowledge about the source data.
\citet{MAHOWALD2024517} state that for learning tasks where language is used in a functional way to accomplish certain goals, such as DORE, self-supervised next-token prediction is not sufficient.
Instead, the model needs to learn task-specific information via specialised fine-tuning to solve such tasks, which are not frequently present in pre-training data and involve task-specific reasoning.
DORE can only be solved by knowing the specific relationship definition provided in the task descriptions, which the model cannot handle if it was not trained on the task.
In summary, the presented results reinforce the observation that fine-tuning and specialised architectures are still needed to solve functional language-based tasks that cannot be solved by pattern matching alone.

%computational complexity
The computational complexity of CoT-decoding is $k$ times higher than regular greedy decoding, however, decoding of the different branches can be done in parallel.
Nonetheless, one should note that only one run of ontology construction is needed, as the ontology can be repeatedly used for other dialogue modelling tasks such as state tracking. %understanding, management and generation. 
It is worth stressing that CoT-decoding only increases inference cost, while training cost remains the same.
Compared to few-shot prompting, which also increases inference cost due to longer input context, there is no additional annotation cost. 

% % % % % % % % % % %
\section{Conclusion}

We propose constrained chain-of-thought~(CoT)-decoding, a new decoding mechanism for dialogue ontology generation~(DORE) in a transfer set-up.
An LLM using a one-shot example from the source data is significantly improved using the proposed constrained CoT-decoding mechanism.
Fine-tuning an LLM on the source data and using constrained CoT-decoding for inference on the target data outperforms a one-shot target data model significantly.

The results warrant further research into DORE in particular, and into eliciting reasoning in LLMs by adapting the decoding mechanism in general.
Moreover, we offer a method for applying LLMs to tasks that are underrepresented in pre-training and where the vanilla LLMs perform poorly. Our method is appealing as it does not necessitate labelling new examples.
Future research directions include explainability of constrained CoT-decoding in fine-tuned LLMs and including CoT-decoding during fine-tuning.

% % % % % % % % % % %
\section{Limitations}

In this work we assume a pipeline approach, however with the raise of LLMs, end-to-end solutions tend to be more accurate. 
We leave the task of jointly extracting dialogue terms and relations for future investigation.
Due to constraints in computational infrastructure, we were not able to run open-source LLMs with the size of ChatGPT, which might be promising however.
We abstained from utilising proprietary models, such as ChatGPT, for increased transparency and reduced risk of training data contamination. 

Furthermore, the need for an annotated source dataset limits the application to low-resource languages and tasks.
The reliance on a specific answer structure for confidence estimation limits application to less structured tasks.

Finally, what we consider the upper bound, which was trained on the target dataset, can be argued to be a low bar too, reaching only an F1 of 37.
This warrants more research on this task also on the same data setting.

% % % % % % % % % % % % % %
\section{Acknowledgements}

RV and BMR are supported by funds from the European Research Council (ERC) provided under the Horizon 2020 research and innovation programme (Grant agreement No.\ STG2018 804636) as part of the DYMO project.
CVN and HL are supported by the Ministry of Culture and Science of North Rhine-Westphalia within the framework of the Lamarr Fellow Network.
MH is supported by funding provided by the Alexander von Humboldt Foundation in the framework of the Sofja Kovalevskaja Award endowed by the Federal Ministry of Education and Research.
Computational infrastructure and support were provided by the Centre for Information and Media Technology at Heinrich Heine University Düsseldorf and Google Cloud.
We want to thank the anonymous reviewers whose comments improved the quality of our paper.

\bibliography{anthology,custom}
% \bibliography{custom}

\appendix

% % % % % % % % % % % % % % % % % % % % % % % % %
\section{Results for Different Relation Types}
\label{sec:appendix:relation_type_results}

See \Cref{tab:sgd_test_relation_type_results} for results on the different ontology relation types.

\begin{table*}[ht]
    \centering
    \small
    \resizebox{0.8\linewidth}{!}{%
    \begin{tabular}{p{5.5cm}|c|ccc}
        \toprule
        \textbf{Approach} & \textbf{Relation Type} & \textbf{F1-Score} & \textbf{Precision} & \textbf{Recall}
        \\
        \midrule
        \multicolumn{5}{c}{
            \textit{One-shot example from MultiWOZ}
        } 
        \\
        % \midrule
        % {One-shot example from MWOZ} & 0.021 & 0.051 & 0.013 \\ 
        \midrule
        \multirow{4}{\linewidth}{\textit{Baseline}: Separate relation prediction} & all & 7.4 & 8.8 & 6.4 \\ 
        & domain-slot & 3.9 & 2.1 & 41.0 \\ 
        & slot-value & 9.4 & 18.8 & 6.3 \\ 
        & value-domain & 8.0 & 12.7 & 5.9 \\ 
        & equivalence & 1.8 & 1.0 & 6.9 \\ 
        \midrule
        \multirow{4}{\linewidth}{\hspace{0.5cm}+ constrained decoding} & all & 8.5 & 5.7 & 17.3 \\ 
        & domain-slot & 1.1 & 0.5 & 50.7 \\ 
        & slot-value & 9.3 & 7.0 & 13.8 \\ 
        & value-domain & 14.4 & 11.5 & 19.5 \\ 
        & equivalence & 1.4 & 0.7 & 16.1 \\ 
        \midrule
        \multirow{4}{\linewidth}{\hspace{0.5cm}+ CoT decoding} & all & 9.2 & 8.8 & 9.6 \\ 
        & domain-slot & 2.5 & 1.3 & 50.7 \\ 
        & slot-value & 16.0 & 18.4 & 14.2 \\ 
        & value-domain & 7.7 & 12.7 & 5.5 \\ 
        & equivalence & 1.8 & 1.0 & 7.3 \\ 
        \midrule
        \multirow{4}{\linewidth}{\hspace{0.5cm}+ constrained CoT decoding} & all & 9.2 & 6.4 & 15.9 \\ 
        & domain-slot & 1.2 & 0.6 & 50.7 \\ 
        & slot-value & 13.4 & 10.3 & 19.2 \\ 
        & value-domain & 12.3 & 11.6 & 13.1 \\ 
        & equivalence & 1.2 & 0.7 & 10.3 \\ 
        \midrule
        \multicolumn{5}{c}{\textit{Fine-tuning on MultiWOZ}} \\
        \midrule
        \multirow{4}{\linewidth}{\textit{Baseline}: Fine-tuning on MultiWOZ} & all & 10.8 & 6.7 & 28.4 \\ 
        & domain-slot & 5.9 & 3.2 & 49.3 \\ 
        & slot-value & 6.5 & 3.7 & 25.3 \\ 
        & value-domain & 20.5 & 15.3 & 30.8 \\ 
        & equivalence & 4.3 & 2.5 & 18.8 \\ 
        \midrule
        \multirow{4}{\linewidth}{\hspace{0.5cm}+ constrained decoding} & all & 11.1 & 6.8 & 30.4 \\ 
        & domain-slot & 4.7 & 2.5 & 52.9 \\ 
        & slot-value & 6.8 & 3.9 & 25.5 \\ 
        & value-domain & 19.8 & 13.9 & 34.1 \\ 
        & equivalence & 3.9 & 2.1 & 21.5 \\ 
        \midrule
        \multirow{4}{\linewidth}{\hspace{0.5cm}+ CoT decoding} & all & 9.3 & 6.3 & 17.4 \\ 
        & domain-slot & 3.9 & 2.0 & 49.3 \\ 
        & slot-value & 5.7 & 3.5 & 15.5 \\ 
        & value-domain & 16.7 & 14.8 & 18.9 \\ 
        & equivalence & 2.3 & 4.4 & 1.5 \\ 
        \midrule
        \multirow{4}{\linewidth}{\hspace{0.5cm}+ constrained CoT decoding} & all & 14.4 & 10.0 & 25.8 \\ 
        & domain-slot & 3.2 & 1.6 & 64.9 \\ 
        & slot-value & 12.1 & 9.0 & 18.4 \\ 
        & value-domain & 19.2 & 13.9 & 30.9 \\ 
        & equivalence & 4.7 & 2.8 & 16.7 \\ 
        \midrule
        \multicolumn{5}{c}{\textit{Upper Bounds using SGD Data}} \\
        \midrule
        % {One-shot example from SGD} & 0.016 & 0.067 & 0.009 \\ 
        \multirow{4}{\linewidth}{One-shot example from SGD + separate relation prediction} & all & 12.9 & 10.7 & 16.4 \\ 
        & domain-slot & 3.4 & 1.8 & 46.3 \\ 
        & slot-value & 17.7 & 20.9 & 15.4 \\ 
        & value-domain & 13.4 & 11.0 & 17.1 \\ 
        & equivalence & 2.5 & 1.4 & 9.1 \\ 
        \midrule
        \multirow{4}{\linewidth}{Fine-tuning on SGD} & all & 36.8 & 26.4 & 60.9 \\ 
        & domain-slot & 7.3 & 3.8 & 88.8 \\ 
        & slot-value & 30.6 & 20.7 & 58.9 \\ 
        & value-domain & 53.5 & 46.9 & 62.1 \\ 
        & equivalence & 12.3 & 6.9 & 57.6 \\ 
        \bottomrule
    \end{tabular}%
    }
    \caption{Ontology Relation Prediction Results for each relation type on the SGD test set.}

    \label{tab:sgd_test_relation_type_results}
\end{table*}

\section{Example Completions for One-shot Equivalence Relation Prompt}
\label{sec:appendix:only_eq_examples}

See \Cref{tab:sgd_mwoz_oneshot_eq_qualitative_example} for example completions for the one-shot equivalence prompt.

\begin{table*}[t]
    \small
    \centering
    %\resizebox{0.78\linewidth}{!}{%
    \begin{tabular}{l|p{0.7\linewidth}}
        \toprule
        Approach & Response \\
        \midrule
        
        Baseline Greedy Decoding & 
        **Domain:** Hotel 

        **Slot:** Type of hotel

        **Value:** Price range
    
        - [price range, refers to same concept as, price]
        - [reference number, refers to same concept as, reference]
    
        **Domain:** Price

        **Slot:** Cost per night

        **Value:** \$ 63
    
        - [cost per night, refers to same concept as, price]
    
        **Domain:** Location

        **Slot:** Street address

        **Value:** 135 carlingview drive etobicoke, ontario m9w 5e7, canada
    
        - [street address, refers to same concept as, location] ...
     \\
        \midrule
        
        Constrained Decoding & 
        **Domain:** Hotel

        **Slot:** Type of hotel

        **Value:** Price range

        - [price per night,refers to same concept as,135 carlingview drive etobicoke, ontario m9w 5e7, canada]

        - [best western plus toronto north york hotel,refers to same concept as,best western plus toronto north york hotel \& suites] ... 
     \\
        \midrule
        
        CoT-Decoding & Here are the predicted synonyms or semantically equivalent terms from the provided dialogue:

        **Price range:**

        - [price range, refers to same concept as, price]
        
        **Reference number:**

        - [reference number, refers to same concept as, reference]
        
        **Hotel:**

        - [hotel, refers to same concept as, best western plus toronto north york hotel \& suites]
        
        **Stay:**

        - [stay, refers to same concept as, nights]
        
        **Price:**

        - [price, refers to same concept as, cost]<eos> 
         \\
        \midrule
        Constrained CoT-Decoding & Here are the predicted synonyms or semantically equivalent terms from the provided dialogue:

        **Price range:**

        - [price per night,refers to same concept as,cost]
        
        **Reference number:**

        - [price per night,refers to same concept as,check in date]
        
        **Hotel:**

        - [hotel,refers to same concept as,best western premier toronto airport carlingview hotel]
        
        **Day:**

        - [tuesday next week,refers to same concept as,check in date]<eos>
         \\
        
        \bottomrule
    \end{tabular}%
    %}
    \caption{
        \label{tab:sgd_mwoz_oneshot_eq_qualitative_example}
        MWOZ one-shot only equivalence model example response excerpts for the different encodings on SDG test dialogue 100. 
        For CoT-decoding methods, only the chosen branch is displayed.
    }
\end{table*}

\end{document}